\RequirePackage[l2tabu, orthodox]{nag}
\documentclass[11pt]{article}

\usepackage[T1]{fontenc}

\usepackage{soul} 

\usepackage{garamond}  
\renewcommand{\rmdefault}{ugm}

\usepackage[bitstream-charter]{mathdesign}
\usepackage{amsmath}
\usepackage[scaled=0.92]{PTSans}

\usepackage[
  paper  = letterpaper,
  left   = 1.65in,
  right  = 1.65in,
  top    = 1.0in,
  bottom = 1.0in,
  ]{geometry}

\usepackage[usenames,dvipsnames]{xcolor}
\definecolor{shadecolor}{gray}{0.9}

\usepackage[final,expansion=alltext]{microtype}
\usepackage[english]{babel}
\usepackage[parfill]{parskip}
\usepackage{afterpage}
\usepackage{framed}

{\endMakeFramed}

\DeclareRobustCommand{\parhead}[1]{\textbf{#1}~}

\usepackage{lineno}

\usepackage{ragged2e}

\newcounter{parcount}

\usepackage{graphicx}
\usepackage[labelfont=bf]{caption}

\usepackage{booktabs}
\usepackage{multirow}

\usepackage[algoruled]{algorithm2e}
\usepackage{listings}
\usepackage{fancyvrb}
\fvset{fontsize=\normalsize}

\usepackage{natbib}

\usepackage[acronym,nowarn]{glossaries}
\makeglossaries

\definecolor{tangerine}{rgb}{0.95, 0.52, 0.0}
\definecolor{palebrown}{rgb}{0.6, 0.46, 0.33}
\definecolor{peru}{rgb}{0.8, 0.52, 0.25}
\usepackage[colorlinks,linktoc=all]{hyperref}
\usepackage[all]{hypcap}
\hypersetup{citecolor=peru}
\hypersetup{linkcolor=peru}
\hypersetup{urlcolor=peru}

\usepackage[nameinlink]{cleveref}
\creflabelformat{equation}{#1#2#3}
\crefname{equation}{eq.}{eqs.}  
\Crefname{equation}{Eq.}{Eqs.}

\lstdefinestyle{mystyle}{
    commentstyle=\color{OliveGreen},
    keywordstyle=\color{BurntOrange},
    numberstyle=\tiny\color{black!60},
    stringstyle=\color{MidnightBlue},
    basicstyle=\ttfamily,
    breakatwhitespace=false,
    breaklines=true,
    captionpos=b,
    keepspaces=true,
    numbers=left,
    numbersep=5pt,
    showspaces=false,
    showstringspaces=false,
    showtabs=false,
    tabsize=2
}
\usepackage[colorinlistoftodos,
           prependcaption,
           textsize=small,
           backgroundcolor=yellow,
           linecolor=lightgray,
           bordercolor=lightgray]{todonotes}

\DeclareRobustCommand{\parhead}[1]{\textbf{#1}~}

\usepackage{amsthm}  

\usepackage{bm}

\usepackage[colorinlistoftodos,
           prependcaption,
           textsize=small,
           backgroundcolor=yellow,
           linecolor=lightgray,
           bordercolor=lightgray]{todonotes}

\usepackage{graphicx}
\usepackage{caption}
\usepackage{subcaption}
\usepackage{wrapfig}

\usepackage{booktabs}
\usepackage{arydshln} 
\usepackage{multirow}
\usepackage{nicematrix}

\usepackage{listings}
\usepackage{fancyvrb}
\fvset{fontsize=\normalsize}

\usepackage[nameinlink]{cleveref}
\creflabelformat{equation}{#1#2#3}
\crefname{equation}{eq.}{eqs.}  
\Crefname{equation}{Eq.}{Eqs.}

\usepackage{listings}
\lstdefinestyle{alp_style}{
    commentstyle=\color{OliveGreen},
    numberstyle=\tiny\color{black!60},
    stringstyle=\color{BrickRed},
    basicstyle=\ttfamily\scriptsize,
    breakatwhitespace=false,
    breaklines=true,
    captionpos=b,
    keepspaces=true,
    numbers=none,
    numbersep=5pt,
    showspaces=false,
    showstringspaces=false,
    showtabs=false,
    tabsize=2
}

\usepackage{capt-of}

\usepackage{lipsum}

\theoremstyle{remark}
\newtheorem*{lemma*}{Lemma}

\newcommand{\cmark}{\ding{51}}%
\newcommand{\xmark}{\ding{55}}

\newcommand{\bz}{\bm{z}}
\newcommand{\by}{\bm{y}}
\newcommand{\br}{\bm{r}}
\newcommand{\bx}{\bm{x}}

\newcommand{\bI}{\bm{I}}
\newcommand{\bmu}{\bm{\mu}}
\newcommand{\bepsilon}{\bm{\epsilon}}

\usepackage{authblk}
\usepackage{enumitem} 
\usepackage{algorithmic}
\usepackage{tabularx}
\usepackage{booktabs}
\usepackage{pifont}
\usepackage[T1]{fontenc}
\usepackage{multirow}
\usepackage{threeparttable}
\newcommand{\first}[1]{\textbf{\textcolor[RGB]{0,0,255}{#1}}} 
\newcommand{\second}[1]{\underline{\textbf{\textcolor[RGB]{255,165,0}{#1}}}} 

\title{\textbf{The Alpha-Alternator: Dynamic Adaptation To Varying Noise Levels In Sequences Using The Vendi Score For Improved Robustness and Performance}}

\author[1, 3]{Mohammad R. Rezaei}
\author[2, 3]{Adji Bousso Dieng}
\affil[1]{Institute of Biomedical Engineering, University of Toronto}
\affil[2]{Department of Computer Science, Princeton University}
\affil[3]{\href{https://vertaix.princeton.edu/}{Vertaix}}

\begin{document}
\maketitle

\begin{abstract}
 \noindent Current state-of-the-art dynamical models, such as Mamba, assume the same level of noisiness for all sequence elements, which limits their performance on noisy temporal data. In this paper, we introduce the \textbf{$\alpha$-Alternator}, a novel generative model for time-dependent data that dynamically adapts to the complexity introduced by varying noise levels in sequences. The $\alpha$-Alternator leverages the Vendi Score (VS), a flexible similarity-based diversity metric, to adjust, at each time step $t$, the influence of the sequence element at time $t$ and the latent representation of the dynamics up to that time step on the predicted future dynamics. This influence is captured by a parameter that is learned and shared across all sequences in a given dataset. The sign of this parameter determines the direction of influence. A negative value indicates a noisy dataset, where a sequence element that increases the VS is considered noisy, and the model relies more on the latent history when processing that element. Conversely, when the parameter is positive, a sequence element that increases the VS is considered informative, and the $\alpha$-Alternator relies more on this new input than on the latent history when updating its predicted latent dynamics. The $\alpha$-Alternator is trained using a combination of observation masking and Alternator loss minimization. Masking simulates varying noise levels in sequences, enabling the model to be more robust to these fluctuations and improving its performance in trajectory prediction, imputation, and forecasting. Our experimental results demonstrate that the $\alpha$-Alternator outperforms both Alternators and state-of-the-art state-space models across neural decoding and time-series forecasting benchmarks.\\

 \noindent \textbf{Keywords:} Dynamics, Alternators, Neuroscience, Time Series, Vendi Scoring
 \end{abstract}

\section{Introduction}
\glsresetall

\begin{figure*}[!t]
    \includegraphics[width=\linewidth]{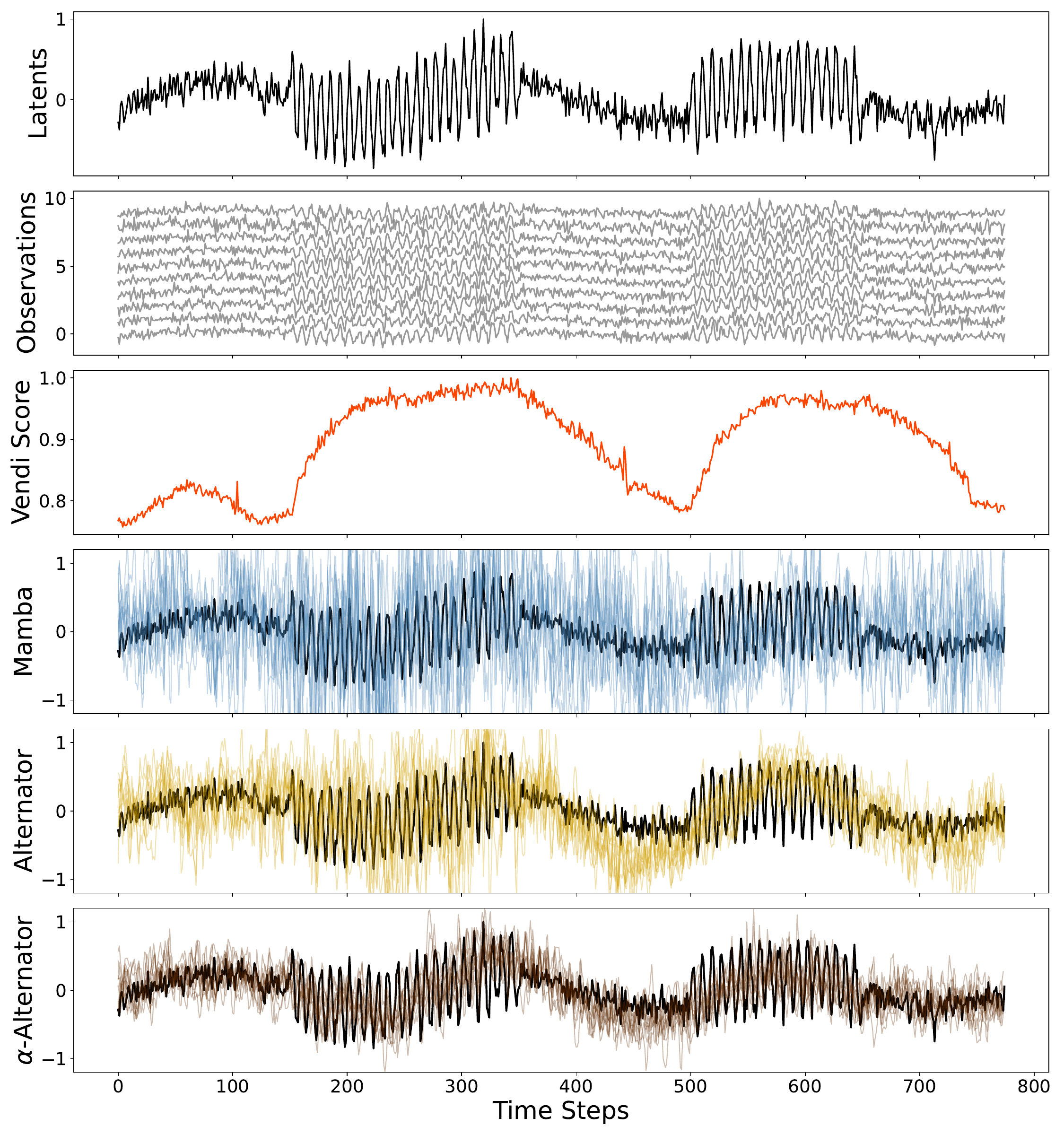}
    \caption{The $\alpha$-Alternator is robust to varying noise levels compared to a Mamba and an Alternator. The Alternator is more robust to noise than the Mamba.}
    \label{fig:motivation}
\end{figure*}

Time-dependent data is central to the natural sciences and engineering disciplines. Modeling such data accurately requires methods that can capture variability both across sequences and within individual sequences. State-space models, such as Mambas, have emerged as a popular framework for sequence modeling~\citep{wang2025mamba,gu2023mamba}. They have demonstrated strong performance in various applications, including speech recognition~\citep{zhang2024mamba} and protein folding~\citep{xu2024protein}. However, Mambas rely on fixed state-space representations that assume smooth transitions across time steps and do not dynamically adjust to noise fluctuations. This is particularly limiting in applications where stochasticity plays a significant role, e.g. neural decoding. 

More recently, Alternators have been introduced as an alternative modeling framework for time-dependent data~\citep{rezaei2024alternators}. Unlike Mambas, which use a structured state-space representation, Alternators explicitly modulate the influence of past and present observations through a gating mechanism, which offers them great flexibility and the ability to capture long-range dependencies. However, despite this flexibility, Alternators still rely on a fixed weighting scheme that does not explicitly account for varying noise levels in the sequence. As a result, they can also struggle in situations where sequence noise fluctuates significantly.

In this work, we introduce the $\alpha$-Alternator, a novel Alternator model that dynamically adjusts its reliance on each sequence element based on its level of noisiness. This mechanism is based on the Vendi Score, a flexible similarity-based diversity metric~\citep{friedman2023vendi}. The $\alpha$-Alternator learns the weight determining its reliance on a given sequence element by applying a sigmoid function to the output of a linear layer, which takes the Vendi Score---computed over two shifted versions of the sequence around that element---as input. The parameter of the linear layer is learned and shared across all sequences in a given dataset, with its sign indicating the direction of influence for each sequence element. When the parameter is negative, sequence elements with large VS values are treated as noisy inputs. As a result, the model places greater emphasis on the latent history when processing these elements. In contrast, when the parameter is positive, sequence elements with high VS are considered informative, and the model relies more on them to update the predicted latent dynamics. This simple mechanism enables the $\alpha$-Alternator to adapt to varying noise levels in sequences and generalize better. We illustrate this behavior in Fig.~\ref{fig:motivation}, where we show the stability of the $\alpha$-Alternator when modeling sequences with varying noise levels across time steps.

The figure illustrates some latent state, simulated using a noisy sine function that incorporates both low-frequency and high-frequency components. The base signal, a sine wave at $2$ Hz, represents the low-frequency component, while higher-frequency components at $60$ Hz are added within two distinct time windows to introduce more complex dynamics. Gaussian noise is then applied to the modulated signal. Ten noisy sequences are drawn as observations by applying random scaling and adding multivariate Gaussian noise, resulting in diverse yet structurally related time series. To quantify the variability of the observations over time, we compute the Vendi Score of the noisy observations using a sliding window of 100 time steps. We then evaluate the performance of the Mamba, the Alternator, and the $\alpha$-Alternator in recovering the latent signal from the ten observations. As shown in the figure, the Mamba struggles with handling highly noisy regions. The Alternator exhibited improved robustness to noise compared to the Mamba, but it still faced challenges in fully adapting to varying noise levels. In contrast, the $\alpha$-Alternator maintains predictive stability even in sequence regions with large amounts of noise.

The performance of the $\alpha$-Alternator was further assessed on neural decoding and time-series forecasting benchmarks. In neural decoding, the model outperformed state-of-the-art baselines in mapping cortical and hippocampal activity to behavioral states. We observed the same thing on time-series forecasting tasks where the $\alpha$-Alternator surpassed Mambas and Alternators among other baselines. 

To understand the contributions of the two key ingredients that make up the $\alpha$-Alternator---the noise adaptation mechanism and the input masking during training---we conducted an ablation study. The findings confirmed that both ingredients are essential for achieving great performance.

\section{Background}
\label{sec:background}

Time-dependent data often exhibits complex dynamics and varying levels of noise across time steps. To effectively model such data, frameworks are needed that can capture the underlying latent dynamics while adapting to input noise. This section outlines the foundations of the $\alpha$-Alternator, described in the next section, which dynamically adjusts its dependency on the current time step or the latent history based on the temporal diversity of the sequence. We begin by describing Alternators, a probabilistic framework for sequence modeling, and then review the Vendi Score, a metric designed to flexibly and accurately quantify diversity. 

\subsection{Alternators}
Consider a sequence \(\bx_{1:T}\). An Alternator models this sequence by coupling it with a sequence of latent variables, \(\bz_{0:T}\), within a joint probability distribution \cite{rezaei2024alternators},
\begin{align}\label{eq:joint}
    p_{\theta, \phi}(\bx_{1:T}, \bz_{0:T}) &= p(\bz_0) \prod_{t=1}^{T} p_{\theta}(\bx_t | \bz_{t-1}) p_{\phi}(\bz_t | \bz_{t-1}, \bx_t)
    .
\end{align}
Here \(p(\bz_0) = \mathcal{N}(0, \bI)\) is a prior distribution over the initial latent variable \(\bz_0\), \(p_{\theta}(\bx_t | \bz_{t-1})\) determines how to generate the sequence elements from the latent state \(\bz_{t-1}\) and \(p_{\phi}(\bz_t | \bz_{t-1}, \bx_t)\) models the evolution of that state over time. Here \(\bz_{t-1}\) acts as a \emph{memory} summarizing the history of the sequence before time \(t\). It is updated dynamically, at each time step, by accounting for both the current state of the memory and the sequence element at time \(t\). This is achieved by defining the mean of \(p_{\phi}(\bz_t | \bz_{t-1}, \bx_t)\) using a gating mechanism,
\begin{align*}
    p_{\phi}(\bz_t | \bz_{t-1}, \bx_t) &= \mathcal{N}\left(\bmu_{z_t}, \sigma_z^2\right), \text{ where }
    \bmu_{z_t} = \sqrt{\alpha_t} \cdot g_{\phi}(\bx_t) + \sqrt{(1 - \alpha_t - \sigma_z^2)}\cdot \bz_{t-1}
    .
\end{align*}
The distribution $p_{\theta}(\bx_t | \bz_{t-1})$ is on the other hand defined as
\begin{align*}
    p_{\theta}(\bx_t | \bz_{t-1}) &= \mathcal{N}\left(\bmu_{x_t},  \sigma_x^2\right) \text{ where }
    \bmu_{x_t} = \sqrt{(1 - \sigma_x^2)}\cdot f_{\theta}(\bz_{t-1})
    .
\end{align*}
Here \(\theta\) and \(\phi\) are parameters of two neural networks and they are learned by minimizing the Alternator loss function
\begin{align}\label{eq:loss-alt}
    \mathcal{L}(\theta, \phi) &= \mathbb{E}_{p(\bx_{1:T}) p_{\theta, \phi}(\bz_{0:T})}\left[\sum_{t=1}^{T} \left\Vert \bz_t - \bmu_{z_t} \right\Vert^2_2 + \frac{D_z \sigma_z^2}{D_x \sigma_x^2} \left\Vert \bx_t - \bmu_{x_t} \right\Vert^2_2 \right],
\end{align}
where \(p(\bx_{1:T})\) is the data distribution and \(p_{\theta, \phi}(\bz_{0:T})\) is the marginal distribution of the latent variables induced by the joint distribution in Eq. \ref{eq:joint}.

\subsection{The Vendi Score}
The Vendi Score (VS) was introduced by \cite{friedman2023vendi} and quantifies the diversity of a collection of elements. Consider a finite set of data points $\left\{\br_1, \dots, \br_n\right\}$. Let $k(\cdot, \cdot)$ denote a positive semi-definite kernel function such that $k(\br_i, \br_i) = 1$ for all $i$. Let $K$ be the corresponding similarity matrix, e.g. $K_{i, j} = k(\br_i, \br_j)$ for all $i, j$. The VS is defined as the exponential of the Shannon entropy of the normalized eigenvalues of $K$,
\begin{align}
    \mathrm{VS}(\left\{\br_1, \dots, \br_n\right\}; k) = \exp\left( - \sum_{i=1}^n \overline{\lambda}_i \log \overline{\lambda}_i \right),
\end{align}
where $\overline{\lambda}_1, \dots, \overline{\lambda}_n$ are the normalized eigenvalues of $K$. The VS is the \emph{effective number} of distinct elements in the dataset and reaches its minimum value of $1$ when all samples are identical (i.e. $k(\br_i, \br_j) = 1$ for all $i \neq j$), and its maximum value of $n$ when all samples are distinct from each other (i.e. $k(\br_i, \br_j) = 0$ for all $i \neq j$).

The VS can be generalized to incorporate sensitivity to rare or common features by using the Renyi entropy of order $q \geq 0$~\citep{pasarkar2024cousins}:
\begin{align}
    \mathrm{VS}_q(\left\{\br_1, \dots, \br_n\right\}; k) = \exp\left( \frac{1}{1 - q} \log \biggl( \sum_{i=1}^n (\overline{\lambda}_i)^q \biggr) \right),
\end{align}
The parameter $q$ controls sensitivity to rarity, with $q = 1$ corresponding to the original Vendi Score. Lower values of $q$ emphasize rare features, while higher values give higher weight to common ones.

The VS's ability to accurately quantify sample diversity without rigid distributional assumptions gives it great flexibility as evidenced by its various applications~\citep{askari2024improving, kannen2024beyond,liu2024diversity,nguyen2024quality, mousavi4924208vsi, pasarkar2023vendi, berns2023towards}. In this paper, we use the VS to measure sequence diversity to define a noise adaptation mechanism for Alternators.

\section{Method}
\label{sec:method}

The $\alpha$-Alternator extends the original Alternator by introducing a mechanism for dynamically adjusting the weighting parameter $\alpha_t$. Consider given $n$ sequences $\bx^{(1)}_{1:T}, \dots, \bx^{(n)}_{1:T}$. The $\alpha$-Alternator first applies a binary mask to these sequences,
\begin{align}
    m_t^{(i)} &\sim \text{Bernoulli}(p_{\text{mask}}) \text{ for all } t \in \left\{1, \dots, T\right\} \text{ and for } i \in \left\{1, \dots, n\right\}\\
    \tilde{\bx}_t^{(i)} &= m_t^{(i)} \cdot \bx_t^{(i)} + (1 - m_t^{(i)}) \cdot \mathbf{0} \text{ for all } t \in \left\{1, \dots, T\right\} \text{ and for } i \in \left\{1, \dots, n\right\} 
\end{align}
where $0 \leq p_{\text{mask}} \leq 1$ is a given masking rate and $\mathbf{0}$ denotes the null vector. At each time step $t$, the $\alpha$-Alternator then computes the \emph{noisiness} of the element $\tilde{\bx}_t^{(i)}$ at that time step using the VS. More specifically, the noisiness of $\tilde{\bx}_t^{(i)}$, which we denote by $\text{VS}_t^{(i)}$, is defined as the VS of two shifted versions of $\tilde{\bx}_{1:T}^{(i)}$, 
\begin{align}
    \text{VS}_t^{(i)} &= \text{VS}\left(\left\{\tilde{\bx}_{t-L:t}^{(i)}, \tilde{\bx}_{t-L+1:t+1}^{(i)}\right\}; k\right)
\end{align}
where $k(\cdot, \cdot)$ is a given positive semi-definite kernel and $L$ is a given window length. The influence of $\tilde{\bx}_t^{(i)}$ is then determined by 
\begin{align}
    \alpha_t^{(i)} = \sigma\left(w\cdot \text{VS}_t^{(i)} + b\right) \cdot (1 - \sigma_z^2 - \epsilon_0),
\end{align}
where $w$ and $b$ are unknown scalar parameters, $\sigma_z^2$ denotes the variance of the distribution $p_{\phi}(\bz_t | \bz_{t-1}, \bx_t)$ as described in Section \ref{sec:background}, and $\epsilon_0$ represents a small constant added to ensure numerical stability. This definition of $\alpha_t$ abides by the constraint $0 \leq \alpha_t < 1 - \sigma_z^2$ that it should satisfy~\citep{rezaei2024alternators}.  

\begin{algorithm}[t]
\DontPrintSemicolon
 Inputs: Data $\bx_{1:T}^{(1:n)}$, batch size $B$, variances $\sigma_x^2$ and $\sigma_z^2$, and masking rate $p_{\text{mask}}$\;
 Initialize model parameters $\theta$, $\phi$, $w$, and $b$\;
 \While{not converged}{
 \For{$b = 1, \dots, B$}{
  Draw initial latent $\bz_0^{(b)} \sim \mathcal{N}(0, I_{D_z})$\;
  \For{$t = 1, \dots, T$}{
      Compute $\bmu_{x_t}^{(b)} = \sqrt{(1 - \sigma_x^2)}\cdot f_{\theta}(\bz_{t-1}^{(b)})$\;
      Sample binary mask $m_t \sim \text{Bernoulli}(p_{mask})$\;
      Apply random masking $\tilde{\mathbf{x}}_t^{(b)} = m_t \cdot \mathbf{x}_t^{(b)} + (1 - m_t) \cdot \mathbf{0}$\;
      Compute adaptive weight $\alpha_t^{(b)} = \sigma\left(w\cdot \text{VS}_t^{(b)} + b\right) \cdot (1 - \sigma_z^2 - \epsilon_0)$\;
      Compute $\bmu_{z_t}^{(b)} = \sqrt{\alpha_t^{(b)}} \cdot g_{\phi}(\Tilde{\bx}_t^{(b)}) + \sqrt{(1 - \alpha_t^{(b)} - \sigma_z^2)}\cdot \bz_{t-1}^{(b)}$\;
      Sample latent $\bz_{t}^{(b)} \sim \mathcal{N}\left(\bmu_{z_t}^{(b)}, \sigma_z^2\right)$
  }}
  Compute loss $\mathcal{L}(\theta, \phi, w, b)$ in Eq. \ref{eq:loss-adaptive}\;
  Backpropagate and update parameters $\theta$, $\phi$, $w$, and $b$\;
 }
 \caption{Sequence modeling with the $\alpha$-Alternator}\label{alg:training}
\end{algorithm}

The $\alpha$-Alternator also modifies the original Alternator loss function described in Eq. \ref{eq:loss-alt}, using the adaptive $\alpha_t^{(i)}$ defined above,
\begin{align}\label{eq:loss-adaptive}
    \mathcal{L}(\theta, \phi, w, b) &= \frac{1}{n}\sum_{i=1}^{n}\mathbb{E}_{p_{\theta, \phi}(\mathbf{z}_{0:T})}
    \left[\sum_{t=1}^{T} \left\Vert \mathbf{z}_t^{(i)} - \mathbf{\mu}_{z_t}^{(i)} \right\Vert^2_2 + \alpha_t^{(i)}  \frac{D_z \sigma_z^2}{D_x \sigma_x^2} \left\Vert \mathbf{x}_t^{(i)} - \mathbf{\mu}_{x_t}^{(i)} \right\Vert^2_2 \right].
\end{align}
where $\bx_t^{(i)}$ is the element at time $t$ of the $i^{\text{th}}$ sequence, before any masking is applied, and $\mathbf{z}_t^{(i)}$ and $\mathbf{\mu}_{z_t}^{(i)}$ are defined as 
\begin{align*}
    \bmu_{z_t}^{(i)} &= \sqrt{\alpha_t^{(i)}} \cdot g_{\phi}(\tilde{\bx}_t^{(i)}) + \sqrt{(1 - \alpha_t^{(i)} - \sigma_z^2)}\cdot \bz_{t-1}^{(i)} \text{ and }  \bz_{t}^{(i)} \sim  \mathcal{N}\left(\bmu_{z_t}^{(i)}, \sigma_z^2\right)
    .
\end{align*}
The role of $\alpha_t^{(i)}$ in determining the influence of the current observation $\bx_t^{(i)}$ when predicting the latent dynamics is apparent. When $\alpha_t^{(i)}$ is high, $\bx_t^{(i)}$ has a strong influence on the prediction of the dynamics. On the other hand, when $\alpha_t^{(i)}$ is low, $\bx_t^{(i)}$ has a low influence on the prediction of the latent dynamics and the model relies more on the latent history $\bz_{t-1}^{(i)}$, which ensures smooth transitions and temporal consistency. In terms of the loss function \(\mathcal{L}(\theta, \phi, w, b)\), $\alpha_t^{(i)}$ affects the reconstruction error terms for both $\bz_{t}^{(i)}$ and $\bx_{t}^{(i)}$. Since $\alpha_t^{(i)}$ scales the contribution of $\bx_{t}^{(i)}$ in the loss, it dynamically adjusts the importance of the observation-based error term relative to the latent transition error term. This enables the model to adaptively shift between short-term reactivity and long-term memory, making it well-suited for handling diverse temporal structures in sequence modeling. Algorithm \ref{alg:training} describes the complete training procedure. 

Once trained, sampling new sequences from the $\alpha$-Alternator is simple, and Algorithm \ref{alg:Sampling} describes the procedure.

\begin{algorithm}[t]
\DontPrintSemicolon
 Inputs: Variances $\sigma_x^2$, $\sigma_z^2$, constant $\epsilon_0$, and learned parameters $\theta$, $\phi$, $w$, $b$\;
  Draw initial latent $\bz_0 \sim \mathcal{N}(0, I_{D_z})$\;
  \For{$t = 1, \dots, T$}{
      Draw noise variables $\bepsilon_{xt}\sim \mathcal{N}(0, I_{D_x})$ and $\bepsilon_{zt}\sim \mathcal{N}(0, I_{D_z})$\;
      Draw $\bx_t= \sqrt{(1 - \sigma_x^2)}\cdot f_{\theta}(\bz_{t-1}) + \sigma_x \cdot \bepsilon_{xt}$\;
      Compute adaptive weight $\alpha_t = \sigma\left(w\cdot \text{VS}_t + b\right) \cdot (1 - \sigma_z^2 - \epsilon_0)$\;
      Draw $\bz_t = \sqrt{\alpha_t} \cdot g_{\phi}({\bx}_t) + \sqrt{(1 - \alpha_t - \sigma_z^2)}\cdot \bz_{t-1} + \sigma_z \cdot \bepsilon_{zt}$
  }
 \caption{Sampling sequences using the $\alpha$-Alternator}\label{alg:Sampling}
\end{algorithm}

\section{Experiments}
\label{sec:empirical}

In this section, we test the $\alpha$-Altenator against strong baselines on neural decoding and time-series forecasting. 

\subsection{Neural Decoding}
Neural decoding is a fundamental challenge in neuroscience, essential for understanding the mechanisms linking brain function and behavior. In neural decoding, neural data are translated into information about variables such as movement, decision-making, perception, or cognitive functions~\citep{donner2009buildup,lin2022predicting,rezaei2018comparison,rezaei2023inferring}.

We use the \(\alpha\)-Alternator to decode neural activities from three distinct experiments, each targeting a different brain region with specialized functional roles. 

In the first experiment, we recorded the 2D velocity of a monkey as it controlled a cursor on a screen, alongside a 21-minute recording from the Motor Cortex (MC), capturing activity from 164 neurons. The motor cortex is responsible for planning and executing voluntary movements, making it a critical region for decoding motion-related neural signals.

The second experiment involved the same monkey performing a similar cursor control task; however, instead of the motor cortex, neural recordings were obtained from the Somatosensory Cortex (SS). This 51-minute recording included 52 neurons. The somatosensory cortex processes sensory inputs such as touch, proprioception, and movement-related feedback, allowing us to explore how sensory-driven neural activity contributes to movement execution and adaptation.

Finally, in the third experiment, we recorded the 2D positions of a rat as it navigated a platform in search of rewards. This session lasted 75 minutes and captured activity from 46 neurons in the hippocampus, a brain region essential for spatial memory and navigation. The hippocampus contains "place cells" that encode location-specific information, providing insights into how neural representations guide movement in a learned environment.

Our objective in using the \(\alpha\)-Alternator on these varied datasets is to demonstrate its effectiveness across brain regions responsible for different cognitive and behavioral roles, such as motor control, sensory integration, and spatial memory/navigation. For more details on datasets, we refer the reader to \cite{glaser2020machine,glaser2018population}. The time horizons for these experiments were divided into 1-second windows for decoding, with a time resolution of $5$ milliseconds. We used the first 70\% of each recording for training and the remaining 30\% as the test set. In this experiment, we define the features as the velocity/position and the observations as the neural activity data.

\parhead{Empirical setup.} For the two networks with parameters $\theta$ and $\phi$ of the $\alpha$-Alternator, we used attention-based models with two layers, each followed by a hidden layer containing $10$ units. We set $\sigma_z = 0.1$ and $\sigma_x = 0.2$. We used a window length $L = 10$ and set $q = 0.2$ when computing the VS with an RBF kernel. 

The model was trained for $500$ epochs on three different datasets: Motor Cortex, Somatosensory, and Hippocampus. We used the Adam optimizer with an initial learning rate of 0.01. Additionally, a cosine annealing learning rate scheduler was applied, with a minimum learning rate of $1e^{-3}$ and $5$ warm-up epochs to stabilize the early training phase. We benchmarked the $\alpha$-Alternator on its ability to accurately predict velocity/position given neural activity against the Alternator, the Mamba, VRNN~\citep{chung2015recurrent}, SRNN~\citep{fraccaro2016sequential}, and Neural ODE or NODE~\citep{chen2018neural}.

\begin{figure*}[t]
\includegraphics[width=\linewidth]{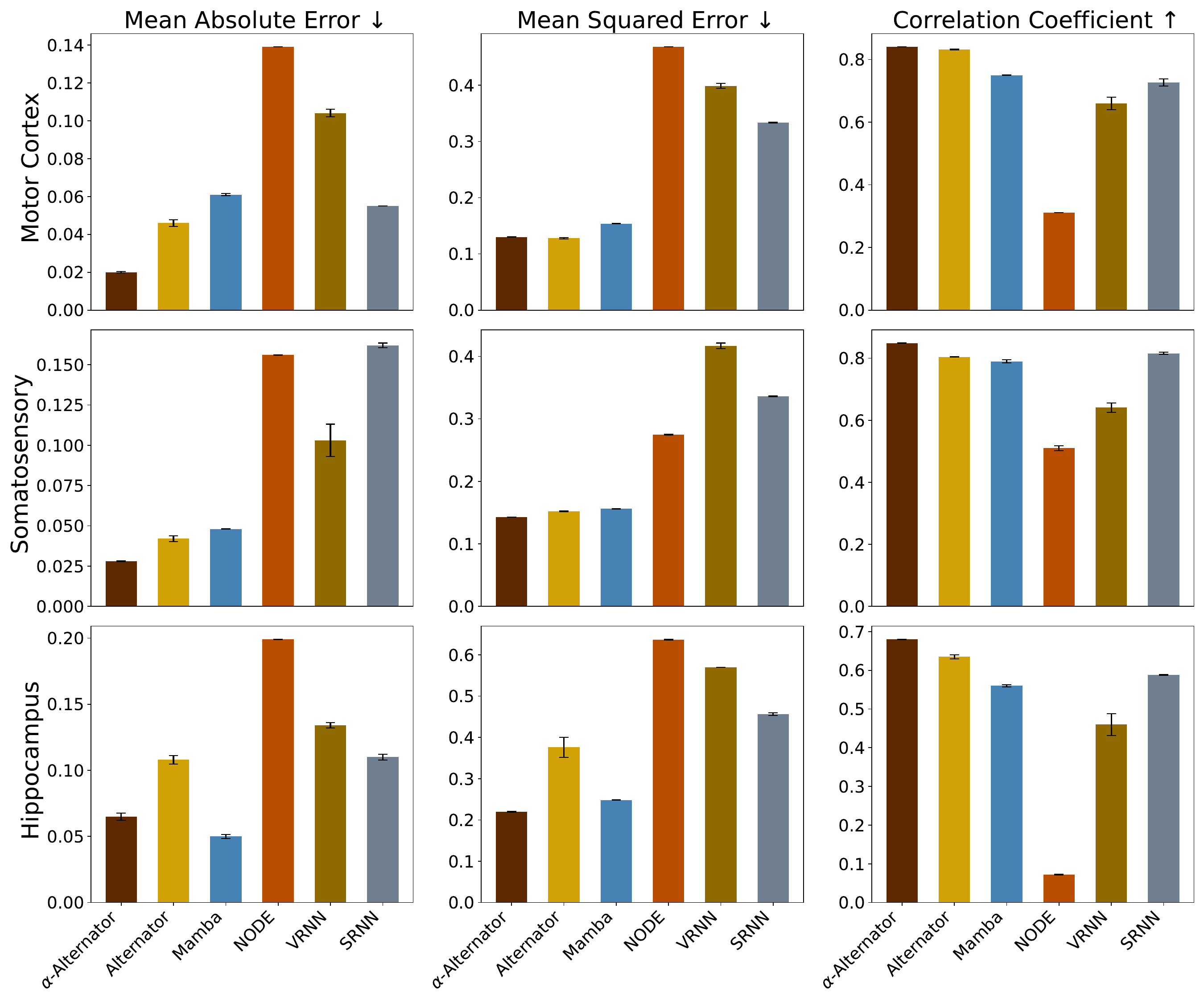}
    \caption{The $\alpha$-Alternator outperforms other models on trajectory prediction in the neural decoding task on all three datasets in terms of MSE and CC. In terms of MAE, the $\alpha$-Alternator outperforms the baselines on all datasets except the Hippocampus dataset, which has lower temporal diversity as shown in Figure \ref{fig:mae-vs-comparision}. 
    }
    \label{fig:neural-result}
\end{figure*}

\begin{figure*}[t]
\centering
\includegraphics[width=.9\linewidth]{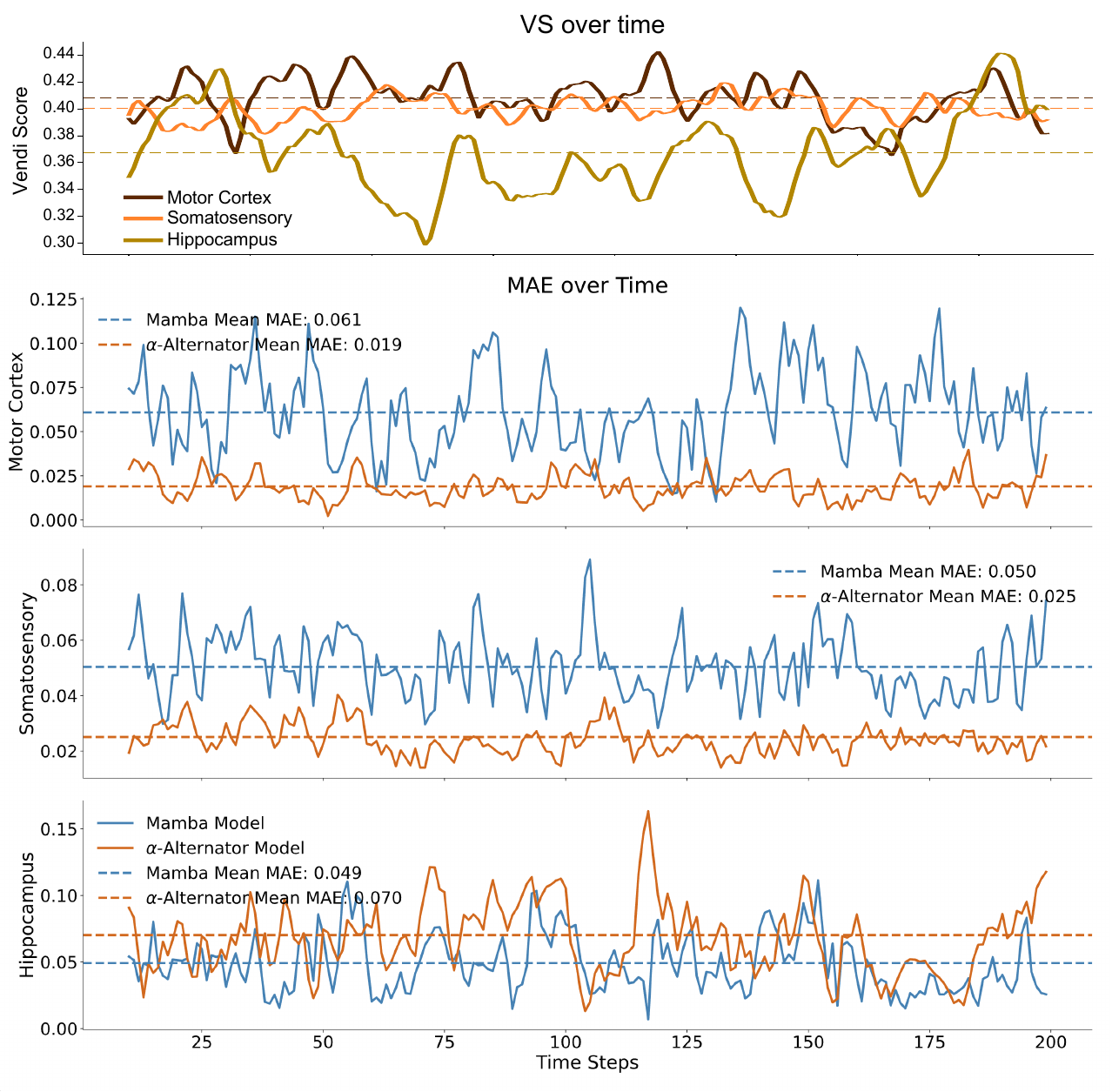}
    \caption{ VS over time for the Motor Cortex, Hippocampus, and Somatosensory Cortex datasets. Lower VS values in the Hippocampus indicate less diverse observations across time steps, leading to a diminished effect of the adaptive mechanism in the $\alpha$-Alternator compared to the Mamba. In contrast, for the Motor Cortex and the Somatosensory datasets, the $\alpha$-Alternator effectively leverages VS-based adaptation, outperforming the Mamba in handling varying noise levels.}
    \label{fig:mae-vs-comparision}
\end{figure*}

\parhead{Results.} As Figure \ref{fig:neural-result} shows, the $\alpha$-Alternator achieves superior performance across all three neural datasets, showing particular strength in handling complex neural decoding tasks. In the Motor Cortex dataset, the model achieves notably lower MAE compared to all baselines, including Mamba, NODE, VRNN, and SRNN, while maintaining the highest CC of approximately $80\%$. This improvement is especially significant given the Motor Cortex's intricate temporal patterns, where the $\alpha$-Alternator's adaptive mechanism appears to capture motion-related neural dynamics more effectively than traditional approaches.

The superiority of the $\alpha$-Alternator extends to the Somatosensory dataset, where it maintains consistently lower error rates across both MAE and MSE metrics. While the baselines, particularly the Mamba and the NODE, show competitive performance in terms of CC, the $\alpha$-Alternator achieves better overall accuracy. This suggests enhanced capability in processing complex somatosensory inputs, where precise temporal relationships are crucial for accurate decoding.

In the Hippocampus dataset, the $\alpha$-Alternator outperforms most of the baselines across multiple performance metrics. It achieves significantly lower MSE values while maintaining the highest CC among all tested models. However, the $\alpha$-Alternator does not surpass the Mamba in terms of MAE on this dataset, despite its advantages in MSE and CC. As shown in Fig.~\ref{fig:mae-vs-comparision}, the average temporal diversity (as measured by VS) in the Hippocampus dataset is lower compared to the Motor Cortex and Somatosensory datasets. This suggests that the observations in the Hippocampus dataset are less diverse over time, which may reduce the effectiveness of the $\alpha$-Alternator's adaptive weighting mechanism.

\parhead{Ablation study.} Our ablation study demonstrates the significant benefits of combining the noise adaptation mechanism, i.e. adaptive $\alpha_t$, with masking across three neural datasets. The experimental results, shown in Table~\ref{tab:Neural_abl_performance}, reveal consistent performance improvements when both components are utilized together.

\begin{table}
\centering
\caption{Ablation study. The MAE, MSE, and CC between the true and predicted trajectories in the neural decoding task on three different datasets are better when using the two ingredients that make up the $\alpha$-Alternator, the noise adaptation mechanism for $\alpha_t$ and the observation masking for training.}
\begin{tabular}{lccccc} 
\hline
Dataset & Adaptive $\alpha_t$? & Masking? & MAE$\downarrow$ & MSE$\downarrow$ & CC$\uparrow$  \\
\hline
\multirow{4}{*}{Motor Cortex} & \xmark & \xmark & 0.041 & {0.130} & 0.837\\
& \xmark & \cmark & 0.046 & \textbf{0.128} & 0.832 \\
& \cmark & \xmark & 0.057 & 0.158 & 0.796 \\
& \cmark & \cmark & \textbf{0.023} & 0.131 & \textbf{0.841} \\\hline
\multirow{4}{*}{Somatosensory} & \xmark & \xmark & 0.042 & 0.152 & 0.804 \\
& \xmark & \cmark & 0.038 & 0.147 & 0.825 \\
& \cmark & \xmark & 0.042 & 0.179 & 0.771 \\
& \cmark & \cmark & \textbf{0.028} & \textbf{0.143} & \textbf{0.849}\\\hline
\multirow{4}{*}{Hippocampus} & \xmark & \xmark & 0.108 & 0.376 & 0.635\\
& \xmark & \cmark & 0.081 & 0.332 & 0.651 \\
& \cmark & \xmark & 0.067 & 0.225 & 0.671 \\
& \cmark & \cmark & \textbf{0.065} & \textbf{0.222} & \textbf{0.681} \\\hline
\end{tabular}
\label{tab:Neural_abl_performance}
\end{table}

In the Motor Cortex dataset, the combination of adaptive $\alpha_t$ and masking achieved the best performance with an MAE of $0.023$ and CC of $0.841$, representing a $43.9\%$ reduction in MAE compared to the baseline model (no adaptive $\alpha_t$, no masking). While using masking alone showed modest improvements in MSE, the full model's superior performance in MAE and CC highlights the synergistic effect of combining both ingredients.

The Somatosensory dataset exhibited similar trends, with the complete model achieving optimal results across all metrics. The improvement is particularly noteworthy compared to using either component in isolation, where masking alone or adaptive $\alpha_t$ alone showed limited benefits. The full model demonstrated a 33.3\% reduction in MAE from the baseline configuration.

Most notably, the Hippocampus dataset showcased the strongest complementary effects between adaptive $\alpha_t$ and masking. The complete model achieved the best performance across all metrics, representing a substantial 39.8\% improvement in MAE compared to the baseline configuration. Interestingly, both adaptive $\alpha_t$ and masking showed individual benefits on this dataset, but their combination led to better results. 

These results consistently demonstrate that while each component offers certain advantages independently, their combination produces the most robust and accurate predictions across different neural regions. 

\begin{figure*}[t]
\includegraphics[width=\linewidth]{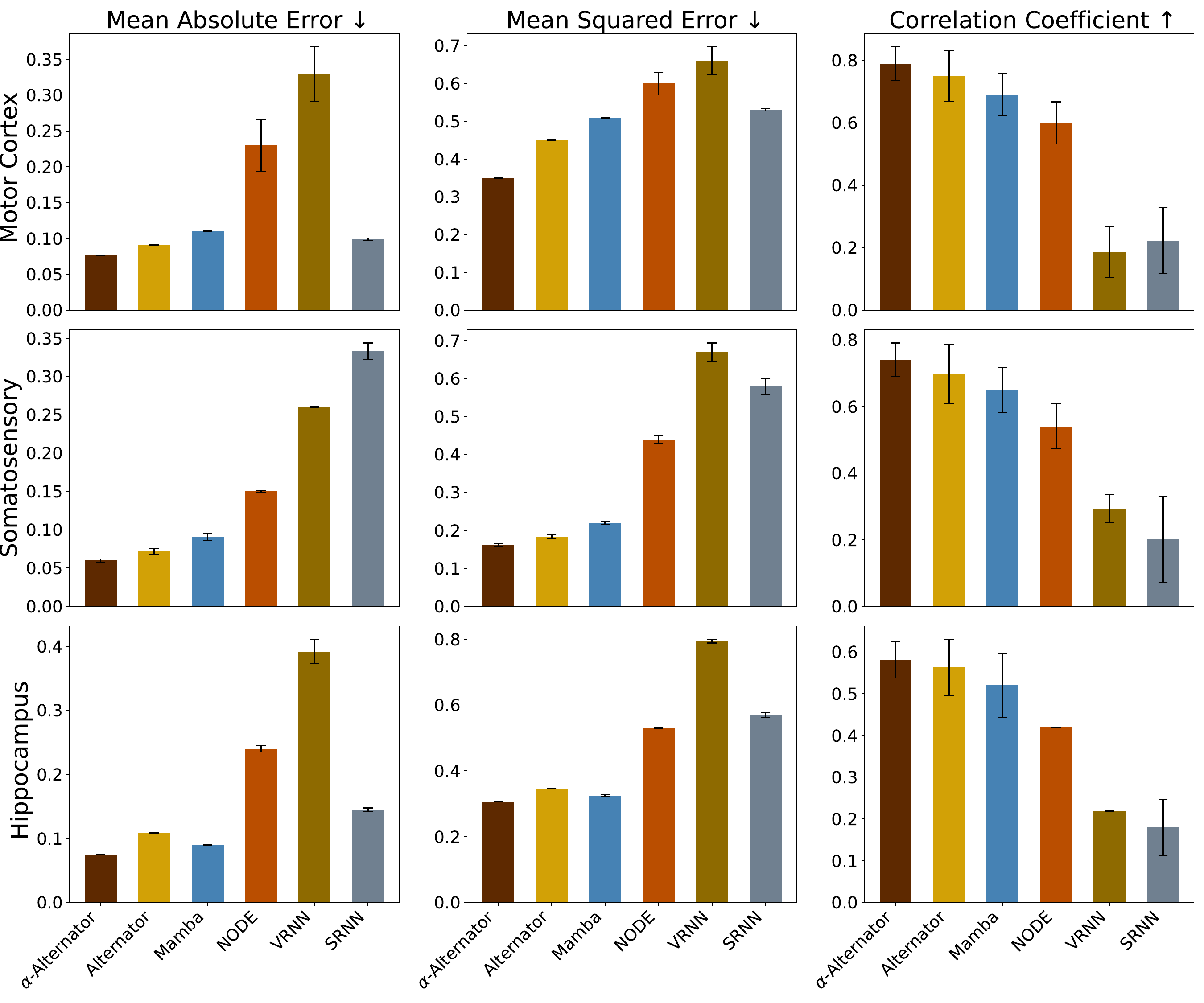}
    \caption{Comparison of performance on neural imputation across different brain regions. The $\alpha$-Alternator consistently outperforms the baselines in imputing missing values across Motor Cortex, Somatosensory, and Hippocampus datasets. Results are averaged across missing value rates ranging from 10\% to 95\%, with performance measured using MAE, MSE, and CC. Vertical bars indicate standard errors across different missing value rates. The $\alpha$-Alternator achieves notably lower errors and higher CCs across all three neural regions, with particularly strong performance in the complex Hippocampus dataset.}
    \label{fig:neural-result-imput}
\end{figure*}

\parhead{Missing value imputation.} As shown in Figure \ref{fig:neural-result-imput}, the $\alpha$-Alternator demonstrates strong robustness in handling missing values, consistently outperforming other models in imputation across neural datasets, even under extreme missing rates ranging from $10\%$ to $95\%$. The $\alpha$-Alternator achieves a lower MAE of approximately 0.06 for the Motor Cortex dataset, surpassing the Mamba model (MAE $\approx$ $0.10$) and showing a particularly notable improvement over NODE and VRNN, both of which have MAE values exceeding $0.20$. The model also excels in MSE performance, maintaining consistently lower values around $0.35$, whereas baseline models, including the Mamba, exhibit higher variability and error rates exceeding $0.5$. Moreover, the $\alpha$-Alternator sustains a high CC of approximately $0.78$, substantially outperforming other models even under challenging missing value conditions. Similar trends are observed in the other datasets.

For the Somatosensory dataset, the $\alpha$-Alternator demonstrates even greater advantages in imputation performance. The model consistently achieves the lowest MAE (approximately $0.06$). Furthermore, its improvements in CC are especially notable, reaching values close to $0.75$, while competing models struggle to maintain reliable correlations under high missing value rates, with the Mamba, for instance, achieving a CC of only around $0.65$.

The Hippocampus dataset poses the most challenging imputation scenario due to its complex spatiotemporal dependencies, yet the $\alpha$-Alternator exhibits good imputation performance on this dataset. The model consistently achieves a lower MSE of approximately $0.07$, outperforming the Mamba (MSE = $0.1$) and the other baselines. The narrow standard error bars of the $\alpha$-Alternator across all metrics indicate stable predictive performance across varying missing value rates, suggesting that the model's adaptive mechanism effectively captures the intricate patterns of hippocampal activity, even under substantial missing data settings. 

The results of the missing value imputation task highlight the robust imputation capabilities of the $\alpha$-Alternator, which excels even in challenging scenarios with high proportions of missing values.

\subsection{Time-series forecasting}
\begin{table*}[t]
  \caption{Forecasting results for the $\alpha$-Alternator and several strong baselines on the Electricity, Exchange, Weather, and Solar-Energy datasets.
  The lookback length $L$ is set to 96 and the forecast length $T$ is set to 96, 192, 336, 720. \first{blue} indicates the best performance while \second{orange} indicates second-best performance.}
  \label{tab:results_forecast}
  \renewcommand{\arraystretch}{0.9}
  \centering
  \resizebox{\textwidth}{!}{
  \begin{threeparttable}
  \begin{small}
  \setlength{\tabcolsep}{2.6pt}
  \begin{tabular}{c|c|cc|cc|cc|cc|cc|cc|cc|cc|cc|cc}
    \toprule
    \multicolumn{2}{c|}{Models} & \multicolumn{2}{c|}{$\alpha-$Alternator} & \multicolumn{2}{c|}{S-Mamba} & \multicolumn{2}{c}{iTransformer} & \multicolumn{2}{c}{\textbf{Alternator}} & \multicolumn{2}{c}{Crossformer} & \multicolumn{2}{c}{TiDE} & \multicolumn{2}{c}{TimesNet} & \multicolumn{2}{c}{DLinear} & \multicolumn{2}{c}{FEDformer} &  \multicolumn{2}{c}{Autoformer} \\
    \cmidrule(lr){1-2}\cmidrule(lr){3-4}\cmidrule(lr){5-6}\cmidrule(lr){7-8}\cmidrule(lr){9-10}\cmidrule(lr){11-12}\cmidrule(lr){13-14}\cmidrule(lr){15-16}\cmidrule(lr){17-18}\cmidrule(lr){19-20}\cmidrule(lr){21-22}  
    \multicolumn{2}{c|}{Metric} & MSE & MAE & MSE & MAE & MSE & MAE & MSE & MAE & MSE & MAE & MSE & MAE & MSE & MAE & MSE & MAE & MSE & MAE & MSE & MAE  \\
    \toprule
    \multirow{5}{*}{\rotatebox{90}{Electricity}} 
    & 96  & \second{0.142} & \second{0.238} & \first{0.139} & \first{0.235} & {0.148} & {0.240} & 0.223 & 0.318 & 0.219 & 0.314 & 0.237 & 0.329 & {0.168} & 0.272 & 0.197 & 0.282 & 0.193 & 0.308 & 0.201 & 0.317\\
    
    & 192 & \first{0.157} & \first{0.252} & \second{0.159} & {0.255} & {0.162} & \second{0.253} & \second{0.162} & 0.256 & 0.231 & 0.322 & 0.236 & 0.330 & 0.184 & 0.289 & 0.196 & 0.285 & 0.201 & 0.315 & 0.222 & 0.334\\
    
    & 336 & \first{0.169} & \first{0.266} & {0.176} & {0.272} & {0.178} & \second{0.269} & \second{0.175} & {0.270} & 0.246 & 0.337 & 0.249 & 0.344 & 0.198 & 0.300 & 0.209 & 0.301 & 0.214 & 0.329 & 0.231 & 0.338\\
    
    & 720 & \first{0.193} & \first{0.289} & \second{0.204} & {0.298} & 0.225 & {0.317} & {0.208} & \second{0.297} & 0.280 & 0.363 & 0.284 & 0.373 & 0.220 & 0.320 & 0.245 & 0.333 & 0.246 & 0.355 & 0.254 & 0.361\\
    \cmidrule(lr){2-22}
    
    & Avg & \first{0.165} & \first{0.259} & \second{0.170} & \second{0.265} & {0.178} & {0.270} & 0.192 & 0.285 & 0.244 & 0.334 & 0.251 & 0.344 & 0.192 & 0.295 & 0.212 & 0.300 & 0.214 & 0.327 & 0.227 & 0.338\\
    \midrule
    
    \multirow{5}{*}{\rotatebox{90}{Exchange}} 
    & 96  & \first{0.086} & \second{0.207} & \first{0.086} & \first{0.206} & \first{0.086} & \first{0.206} & 0.093 & 0.216 & 0.256 & 0.367 & 0.094 & 0.218 & 0.107 & 0.234 & \second{0.088} & 0.218 & 0.148 & 0.278 & 0.197 & 0.323\\
    
    & 192 & 0.178 & \second{0.300} & {0.182} & {0.304} & \second{0.177} & \first{0.299} & {0.183} &{0.306} & 0.470 & 0.509 & {0.184} & {0.307} & 0.226 & 0.344 & \first{0.176} & 0.315 & 0.271 & 0.315 & 0.300 & 0.369\\
    
    & 336 & 0.332 & \first{0.417} & \second{0.328} & \second{0.415} & {0.331} & \first{0.417} & 0.336 & 0.420 & 1.268 & 0.883 & {0.349} & {0.431} & 0.367 & 0.448 & \first{0.313} & 0.427 & 0.460 & 0.427 & 0.509 & 0.524\\
    
    & 720 & \first{0.836} & \first{0.689} & {0.867} & {0.703} & {0.847} & \second{0.691} & 0.855 & 0.698 & 1.767 & 1.068 & {0.852} & {0.698} & 0.964 & 0.746 & \second{0.839} & {0.695} & 1.195 & {0.695} & 1.447 & 0.941\\
    \cmidrule(lr){2-22}
    & Avg & \second{0.358} & \first{0.403} & {0.367} & \second{0.408} & {0.360} & \first{0.403} & {0.366} & {0.410} & 0.940 & 0.707 & {0.370} & {0.413} & 0.416 & 0.443 & \first{0.354} & 0.414 & 0.519 & 0.429 & 0.613 & 0.539\\
    \midrule
    
    \multirow{5}{*}{\rotatebox{90}{Weather}} 
    & 96  & \second{0.165} & \first{0.207} & \second{0.165} & \second{0.210} & 0.174 & {0.214} & 0.175 & 0.215 & \first{0.158} & 0.230 & 0.202 & 0.261 & 0.172 & 0.220 & 0.196 & 0.255 & 0.217 & 0.296 & 0.266 & 0.336\\
    
    & 192 & 0.228 & 0.260 & \second{0.214} & \first{0.252} & 0.221 & \second{0.254} & {0.222} & 0.258 & \first{0.206} & 0.277 & 0.242 & 0.298 & 0.219 & 0.261 & 0.237 & 0.296 & 0.276 & 0.336 & 0.307 & 0.367\\
    
    & 336 & \first{0.272} & \first{0.295} & \second{0.274} & {0.297} & {0.278} & \second{0.296} & 0.284 & 0.301 & \first{0.272} & 0.335 & 0.287 & 0.335 & {0.280} & 0.306 & 0.283 & 0.335 & 0.339 & 0.380 & 0.359 & 0.395\\
    
    & 720 & 0.351 & 0.348 & \second{0.350} & \first{0.345} & {0.358} & \second{0.347} & 0.362 & 0.353 & {0.398} & 0.418 & {0.351} & {0.386} & 0.365 & 0.359 & \first{0.345} & {0.381} & 0.403 & 0.428 & 0.419 & 0.428\\
    \cmidrule(lr){2-22}
    
    & Avg & \second{0.254} & \second{0.278} & \first{0.251} & \first{0.276} & {0.258} & \second{0.278} & 0.262 & 0.281 & {0.259} & 0.315 & 0.271 & 0.320 & {0.259} & 0.287 & 0.265 & 0.317 & 0.309 & 0.360 & 0.338 & 0.382\\
    \midrule
    
    \multirow{5}{*}{\rotatebox{90}{Solar-Energy}} 
    & 96  & \first{0.202} & 0.242 & \second{0.205} & {0.244} & \second{0.203} & \first{0.237} & \second{0.205} & \second{0.238} & 0.310 & 0.331 & 0.312 & 0.399 & 0.250 & 0.292 & 0.290 & 0.378 & 0.242 & 0.342 & 0.884 & 0.711  \\
    
    & 192 & \second{0.234} & \first{0.261} & {0.237} & {0.270} & \first{0.233} & \first{0.261} & {0.239} & \second{0.264} & 0.0 & 0.725 & 0.339 & 0.416 & 0.296 & 0.318 & 0.320 & 0.398 & 0.285 & 0.380 & 0.834 & 0.692  \\
    
    & 336 & \first{0.248} & \second{0.276} & {0.258} & {0.288} & \first{0.248} & \first{0.273} & \second{0.250} &\second{0.276} & 0.750 & 0.735 & 0.368 & 0.430 & 0.319 & 0.330 & 0.353 & 0.415 & 0.282 & 0.376 & 0.941 & 0.723  \\
    
    & 720 & \second{0.250} & \second{0.277} & {0.260} & {0.288} & \first{0.249} & \first{0.275} & 0.253 & 0.279 & 0.769 & 0.765 & 0.370 & 0.425 & {0.338} & {0.337} & 0.356 & 0.413 & {0.357} & 0.427 & 0.882 & 0.717  \\
    \cmidrule(lr){2-22}
    
    & Avg & \second{0.234} & \second{0.264} & {0.240} & {0.273} & \first{0.233} & \first{0.262} & {0.236} & \second{0.264} & 0.641 & 0.639 & 0.347 & 0.417 & 0.301 & 0.319 & 0.330 & 0.401 & {0.291} & 0.381 & 0.885 & 0.711\\
    \bottomrule
  \end{tabular}
  \end{small}
  \end{threeparttable}
  }
\end{table*}

We evaluated the effectiveness of the $\alpha$-Alternator across four time-series forecasting benchmarks, each presenting unique challenges. The forecasting performance of the $\alpha$-Alternator and the baselines is measured using MAE and MSE across four different lookback lengths $L$. Table \ref{tab:results_forecast} summarizes the results. The best and second-best models are highlighted in blue and orange, respectively. 

The \textbf{Electricity} dataset, which records hourly consumption patterns of 321 customers from 2012 to 2014, showcases the $\alpha$-Alternator's superior performance in handling multivariate periodic data. Notably, the $\alpha$-Alternator achieved the best performance with an average MSE of $0.165$ and MAE of $0.259$, outperforming both the Alternator and other state-of-the-art models. Compared to S-Mamba (MSE: 0.170, MAE: 0.265), the $\alpha$-Alternator demonstrated a notable improvement across all forecasting horizons, particularly excelling in longer forecasting windows. In the challenging 720-hour forecast length, the $\alpha$-Alternator maintained a lower MSE (0.193) and MAE (0.289) compared to S-Mamba (MSE: 0.204, MAE: 0.298), confirming its robustness in long-term forecasting.

The \textbf{Solar-Energy} dataset comprises 10-minute interval data from 137 photovoltaic plants. While iTransformer showed slightly better performance in terms of average metrics (MSE: 0.233, MAE: 0.262), the $\alpha$-Alternator achieved similar results (MSE: 0.234, MAE: 0.264) and outperformed other models including S-Mamba with an average MSE of 0.240 and MAE of 0.273.

In the \textbf{Exchange} dataset, which presents the complex challenge of forecasting aperiodic daily exchange rates across eight countries from 1990 to 2016, the $\alpha$-Alternator also outperformed the strongest baselines. The model achieved the best MAE of 0.403 and second-best MSE of 0.358 in average performance, showing particular strength in long-term forecasting where it secured the best performance in the 720-day horizon (MSE: 0.836, MAE: 0.689) setting, surpassing S-Mamba (MSE: 0.867, MAE: 0.703), highlighting its effectiveness in handling complex and volatile financial sequences.

For the \textbf{Weather} dataset, the $\alpha$-Alternator achieved overall strong performance (MSE: 0.254, MAE: 0.278), closely following S-Mamba (MSE: 0.251, MAE: 0.276).

Overall, the $\alpha$-Alternator emerges as the top-performing model for these challenging time-series forecasting benchmarks, ranking first or second in most scenarios. 

\section{Related Work}
\label{sec:related}

\parhead{State-Space Models.} State-space models (SSMs) have emerged as a popular framework for modeling time-dependent data across various domains~\citep{gu2023mamba, rezaei2022direct,rezaei2021real,auger2021guide,rangapuram2018deep}. Recent advancements include the Mamba architecture~\citep{gu2023mamba}, which employs a selective state space mechanism defined by 
\begin{align*}
    \mathbf{h}_t &= \text{SSM}(\bx_t, \mathbf{h}_{t-1}), \quad \by_t = \text{Linear}(\mathbf{h}_t)
\end{align*}
where $\mathbf{h}_t$ represents the hidden state, $\bx_t$ is the input, and $\by_t$ is the output at time $t$. In contrast, the $\alpha$-Alternator employs a dynamic state transition, see Algorithm \ref{alg:Sampling}, where $\mathbf{z}_t$ serves an analogous role to Mamba's $\mathbf{h}_t$ but with explicit control over state transitions through $\alpha_t$. While Mamba has demonstrated success in applications from speech recognition~\citep{zhang2024mamba} to protein folding~\citep{xu2024protein}, its architecture requires high-dimensional hidden states $\mathbf{h}_t \in \mathbb{R}^d$ that have the same dimensionality as the data. The $\alpha$-Alternator addresses this limitation by operating in a lower-dimensional latent space $\mathbf{z}_t \in \mathbb{R}^{d_z}$ where $d_z \ll d$, while incorporating the adaptive weighting mechanism to balance between observation influence and state persistence. The lower dimensional state of the $\alpha$-Alternator ($d_z \ll d$) yields reduced computational complexity while the adaptive weighting mechanism is particularly beneficial for stochastic processes like neural recordings where noise characteristics vary significantly over time.

\parhead{Alternators.} The Alternator framework~\citep{rezaei2024alternators} represents a significant departure from traditional SSMs by introducing a dual-network architecture that alternates between producing observations and low-dimensional latent variables over time. The parameters of these two networks are learned by minimizing a cross entropy criterion over the resulting trajectories~\citep{rezaei2024alternators}. This approach has demonstrated superior performance compared to established methods such as Neural ODEs~\citep{chen2018neural}, dynamical VAEs such as VRNNs~\citep{gregor2014deep}, and diffusion models~\citep{dutordoir2022neural, lin2023diffusion} across various sequence modeling tasks. However, the Alternator uses a fixed weighting parameter $\alpha$ when defining the mean of the latent states, which is limiting. The $\alpha$-Alternator extends this framework by letting $\alpha$ vary across time steps using the Vendi Score to automatically adjust its reliance on observations versus latent history. The $\alpha$-Alternator maintains the computational efficiency of the original Alternator while providing greater robustness to temporal variations in sequence noise. Furthermore, the $\alpha$-Alternator's masking strategy during training strengthens its ability to handle missing or corrupted data, a common challenge in real-world applications such as neural decoding and time-series forecasting.

\section{Conclusion}
\label{sec:conclusion}

In this work, we introduced the $\alpha$-Alternator, a novel sequence model designed to overcome the limitations of Alternators and existing state-space models by dynamically adapting to varying noise levels in sequences. The $\alpha$-Alternator leverages the Vendi Score to determine the influence of sequence elements on the prediction of the latent dynamics through a gating mechanism. This same influence score is used to weigh the data reconstruction term in the Alternator loss. The model is trained by masking sequence elements at random during training to simulate varying noise levels. We demonstrate the effectiveness of the $\alpha$-Alternator through an extensive empirical study on neural decoding and time-series forecasting tasks, where we show that it consistently outperforms several state-of-the-art sequence models, including Mambas and Alternators.

\parhead{Limitations and future work.} One limitation of the $\alpha$-Alternator is its assumption of fixed variance---$\sigma_z^2$ and $\sigma_x^2$ are kept constant throughout the sequence---for the distributions over the latent dynamics and observations. Future work will focus on modeling these variances for even greater flexibility.

\section*{Acknowledgements}
Adji Bousso Dieng acknowledges support from the National Science Foundation, Office of Advanced Cyberinfrastructure (OAC): \#2118201. She also acknowledges Schmidt Sciences for their AI2050 Early Career Fellowship.

\section*{Dedication}
This paper is dedicated to \href{https://en.wikipedia.org/wiki/Patrice_Lumumba}{Patrice Émery Lumumba}.

\bibliographystyle{apa}
\bibliography{arxiv}

\begin{thebibliography}{}

\bibitem[\protect\astroncite{Askari~Hemmat et~al.}{2024}]{askari2024improving}
Askari~Hemmat, R., Hall, M., Sun, A., Ross, C., Drozdzal, M., and Romero-Soriano, A. (2024).
\newblock Improving geo-diversity of generated images with contextualized vendi score guidance.
\newblock In {\em European Conference on Computer Vision}, pages 213--229. Springer.

\bibitem[\protect\astroncite{Auger-M{\'e}th{\'e} et~al.}{2021}]{auger2021guide}
Auger-M{\'e}th{\'e}, M., Newman, K., Cole, D., Empacher, F., Gryba, R., King, A.~A., Leos-Barajas, V., Mills~Flemming, J., Nielsen, A., Petris, G., et~al. (2021).
\newblock A guide to state--space modeling of ecological time series.
\newblock {\em Ecological Monographs}, 91(4):e01470.

\bibitem[\protect\astroncite{Berns et~al.}{2023}]{berns2023towards}
Berns, S., Colton, S., and Guckelsberger, C. (2023).
\newblock Towards mode balancing of generative models via diversity weights.
\newblock {\em arXiv preprint arXiv:2304.11961}.

\bibitem[\protect\astroncite{Chen et~al.}{2018}]{chen2018neural}
Chen, R.~T., Rubanova, Y., Bettencourt, J., and Duvenaud, D.~K. (2018).
\newblock Neural ordinary differential equations.
\newblock {\em Advances in neural information processing systems}, 31.

\bibitem[\protect\astroncite{Chung et~al.}{2015}]{chung2015recurrent}
Chung, J., Kastner, K., Dinh, L., Goel, K., Courville, A.~C., and Bengio, Y. (2015).
\newblock A recurrent latent variable model for sequential data.
\newblock {\em Advances in neural information processing systems}, 28.

\bibitem[\protect\astroncite{Donner et~al.}{2009}]{donner2009buildup}
Donner, T.~H., Siegel, M., Fries, P., and Engel, A.~K. (2009).
\newblock Buildup of choice-predictive activity in human motor cortex during perceptual decision making.
\newblock {\em Current Biology}, 19(18):1581--1585.

\bibitem[\protect\astroncite{Dutordoir et~al.}{2022}]{dutordoir2022neural}
Dutordoir, V., Saul, A., Ghahramani, Z., and Simpson, F. (2022).
\newblock Neural diffusion processes.
\newblock {\em arXiv preprint arXiv:2206.03992}.

\bibitem[\protect\astroncite{Fraccaro et~al.}{2016}]{fraccaro2016sequential}
Fraccaro, M., S{\o}nderby, S.~K., Paquet, U., and Winther, O. (2016).
\newblock Sequential neural models with stochastic layers.
\newblock {\em Advances in neural information processing systems}, 29.

\bibitem[\protect\astroncite{Friedman and Dieng}{2023}]{friedman2023vendi}
Friedman, D. and Dieng, A.~B. (2023).
\newblock {The Vendi Score: A Diversity Evaluation Metric for Machine Learning}.
\newblock {\em Transactions on Machine Learning Research}.

\bibitem[\protect\astroncite{Glaser et~al.}{2020}]{glaser2020machine}
Glaser, J.~I., Benjamin, A.~S., Chowdhury, R.~H., Perich, M.~G., Miller, L.~E., and Kording, K.~P. (2020).
\newblock Machine learning for neural decoding.
\newblock {\em Eneuro}, 7(4).

\bibitem[\protect\astroncite{Glaser et~al.}{2018}]{glaser2018population}
Glaser, J.~I., Perich, M.~G., Ramkumar, P., Miller, L.~E., and Kording, K.~P. (2018).
\newblock Population coding of conditional probability distributions in dorsal premotor cortex.
\newblock {\em Nature communications}, 9(1):1788.

\bibitem[\protect\astroncite{Gregor et~al.}{2014}]{gregor2014deep}
Gregor, K., Danihelka, I., Mnih, A., Blundell, C., and Wierstra, D. (2014).
\newblock Deep autoregressive networks.
\newblock In {\em International Conference on Machine Learning}, pages 1242--1250. PMLR.

\bibitem[\protect\astroncite{Gu and Dao}{2023}]{gu2023mamba}
Gu, A. and Dao, T. (2023).
\newblock Mamba: Linear-time sequence modeling with selective state spaces.
\newblock {\em arXiv preprint arXiv:2312.00752}.

\bibitem[\protect\astroncite{Kannen et~al.}{2024}]{kannen2024beyond}
Kannen, N., Ahmad, A., Andreetto, M., Prabhakaran, V., Prabhu, U., Dieng, A.~B., Bhattacharyya, P., and Dave, S. (2024).
\newblock Beyond aesthetics: Cultural competence in text-to-image models.
\newblock {\em arXiv preprint arXiv:2407.06863}.

\bibitem[\protect\astroncite{Lin et~al.}{2022}]{lin2022predicting}
Lin, B., Bouneffouf, D., and Cecchi, G. (2022).
\newblock Predicting human decision making in psychological tasks with recurrent neural networks.
\newblock {\em PloS one}, 17(5):e0267907.

\bibitem[\protect\astroncite{Lin et~al.}{2023}]{lin2023diffusion}
Lin, L., Li, Z., Li, R., Li, X., and Gao, J. (2023).
\newblock Diffusion models for time series applications: A survey.
\newblock {\em arXiv preprint arXiv:2305.00624}.

\bibitem[\protect\astroncite{Liu et~al.}{2024}]{liu2024diversity}
Liu, T.-W., Nguyen, Q., Dieng, A.~B., and G{\'o}mez-Gualdr{\'o}n, D.~A. (2024).
\newblock Diversity-driven, efficient exploration of a mof design space to optimize mof properties.
\newblock {\em Chemical Science}, 15(45):18903--18919.

\bibitem[\protect\astroncite{Mousavi and Khalili}{}]{mousavi4924208vsi}
Mousavi, M. and Khalili, N.
\newblock Vsi: An interpretable bayesian feature ranking method based on vendi score.
\newblock {\em Available at SSRN 4924208}.

\bibitem[\protect\astroncite{Nguyen and Dieng}{2024}]{nguyen2024quality}
Nguyen, Q. and Dieng, A.~B. (2024).
\newblock {Quality-Weighted Vendi Scores And Their Application To Diverse Experimental Design}.
\newblock In {\em International Conference on Machine Learning}.

\bibitem[\protect\astroncite{Pasarkar et~al.}{2023}]{pasarkar2023vendi}
Pasarkar, A.~P., Bencomo, G.~M., Olsson, S., and Dieng, A.~B. (2023).
\newblock Vendi sampling for molecular simulations: Diversity as a force for faster convergence and better exploration.
\newblock {\em The Journal of chemical physics}, 159(14).

\bibitem[\protect\astroncite{Pasarkar and Dieng}{2024}]{pasarkar2024cousins}
Pasarkar, A.~P. and Dieng, A.~B. (2024).
\newblock {Cousins Of The Vendi Score: A Family Of Similarity-Based Diversity Metrics For Science And Machine Learning}.
\newblock In {\em International Conference on Artificial Intelligence and Statistics}, pages 3808--3816. PMLR.

\bibitem[\protect\astroncite{Rangapuram et~al.}{2018}]{rangapuram2018deep}
Rangapuram, S.~S., Seeger, M.~W., Gasthaus, J., Stella, L., Wang, Y., and Januschowski, T. (2018).
\newblock Deep state space models for time series forecasting.
\newblock {\em Advances in neural information processing systems}, 31.

\bibitem[\protect\astroncite{Rezaei et~al.}{2021}]{rezaei2021real}
Rezaei, M.~R., Arai, K., Frank, L.~M., Eden, U.~T., and Yousefi, A. (2021).
\newblock Real-time point process filter for multidimensional decoding problems using mixture models.
\newblock {\em Journal of neuroscience methods}, 348:109006.

\bibitem[\protect\astroncite{Rezaei and Dieng}{2024}]{rezaei2024alternators}
Rezaei, M.~R. and Dieng, A.~B. (2024).
\newblock Alternators for sequence modeling.
\newblock {\em arXiv preprint arXiv:2405.11848}.

\bibitem[\protect\astroncite{Rezaei et~al.}{2018}]{rezaei2018comparison}
Rezaei, M.~R., Gillespie, A.~K., Guidera, J.~A., Nazari, B., Sadri, S., Frank, L.~M., Eden, U.~T., and Yousefi, A. (2018).
\newblock A comparison study of point-process filter and deep learning performance in estimating rat position using an ensemble of place cells.
\newblock In {\em 2018 40th Annual International Conference of the IEEE Engineering in Medicine and Biology Society (EMBC)}, pages 4732--4735. IEEE.

\bibitem[\protect\astroncite{Rezaei et~al.}{2022}]{rezaei2022direct}
Rezaei, M.~R., Hadjinicolaou, A.~E., Cash, S.~S., Eden, U.~T., and Yousefi, A. (2022).
\newblock Direct discriminative decoder models for analysis of high-dimensional dynamical neural data.
\newblock {\em Neural Computation}, 34(5):1100--1135.

\bibitem[\protect\astroncite{Rezaei et~al.}{2023}]{rezaei2023inferring}
Rezaei, M.~R., Jeoung, H., Gharamani, A., Saha, U., Bhat, V., Popovic, M.~R., Yousefi, A., Chen, R., and Lankarany, M. (2023).
\newblock Inferring cognitive state underlying conflict choices in verbal stroop task using heterogeneous input discriminative-generative decoder model.
\newblock {\em Journal of Neural Engineering}, 20(5):056016.

\bibitem[\protect\astroncite{Wang et~al.}{2025}]{wang2025mamba}
Wang, Z., Kong, F., Feng, S., Wang, M., Yang, X., Zhao, H., Wang, D., and Zhang, Y. (2025).
\newblock Is mamba effective for time series forecasting?
\newblock {\em Neurocomputing}, 619:129178.

\bibitem[\protect\astroncite{Xu et~al.}{2024}]{xu2024protein}
Xu, B., Lu, Y., Inoue, Y., Lee, N., Fu, T., and Chen, J. (2024).
\newblock Protein-mamba: Biological mamba models for protein function prediction.
\newblock {\em arXiv preprint arXiv:2409.14617}.

\bibitem[\protect\astroncite{Zhang et~al.}{2024}]{zhang2024mamba}
Zhang, X., Zhang, Q., Liu, H., Xiao, T., Qian, X., Ahmed, B., Ambikairajah, E., Li, H., and Epps, J. (2024).
\newblock Mamba in speech: Towards an alternative to self-attention.
\newblock {\em arXiv preprint arXiv:2405.12609}.

\end{thebibliography}

\end{document}